\documentclass[letterpaper, 10 pt, conference]{ieeeconf}
\usepackage{amsmath,amssymb,amsfonts}
\usepackage{float}
\usepackage{gensymb}
\usepackage{algorithmic}
\usepackage{graphicx}
\usepackage{textcomp}
\usepackage{url}
\usepackage{tabularx}
\usepackage{xcolor}
\usepackage{siunitx}
\usepackage{bm}
\usepackage{amsmath}
\usepackage{multirow}
\usepackage{ragged2e}
\usepackage{subcaption}
\newcolumntype{L}{>{\raggedright\arraybackslash}X}
\def\BibTeX{{\rm B\kern-.05em{\sc i\kern-.025em b}\kern-.08em
    T\kern-.1667em\lower.7ex\hbox{E}\kern-.125emX}}
    
\IEEEoverridecommandlockouts                              

\usepackage[normalem]{ulem}

\title{\LARGE \bf
Active Inference for Integrated State-Estimation, Control, and Learning.}

\author{Mohamed Baioumy
\and
Paul Duckworth
\and
Bruno Lacerda
\and
Nick Hawes
\thanks{Authors are with the Oxford Robotics Institute, University of Oxford, United Kingdom. For correspondence: \{mohamed, pduckworth, bruno, nickh\}@robots.ox.ac.uk}
}

\begin{document}

\maketitle
\thispagestyle{empty}
\pagestyle{empty}

\newcommand{\DriveFullName}{--}
\newcommand{\realposterior}{R_t}
\newcommand{\generativemodel}{p_t}
\newcommand{\variationaldistribution}{q}
\newcommand{\variationaldistributiontime}{q_t}
\newcommand{\Observation}{o}
\newcommand{\action}{a}
\newcommand{\state}{s}
\newcommand{\latentstate}{s^*}
\newcommand{\laplaceF}{F_L}
\newcommand{\laplacemean}{\mu}
\newcommand{\policy}{\pi}
\newcommand{\nameofapproach}{FAIC}

\begin{abstract}
This work presents an approach for control, state-estimation and learning model (hyper)parameters for robotic manipulators. It is based on the active inference framework, prominent in computational neuroscience as a theory of the brain, where behaviour arises from minimizing variational free-energy. First,  we show there is a direct relationship between active inference controllers, and classic methods such as PID control.
We demonstrate its application for adaptive and robust behaviour of a robotic manipulator that rivals state-of-the-art. Additionally, we show that by learning specific hyperparameters, our approach can deal with unmodeled dynamics, damps oscillations, and is robust against poor initial parameters. The approach is validated on the `Franka Emika Panda' 7 DoF manipulator. Finally, we highlight limitations of active inference controllers for robotic systems.
\end{abstract}

\begin{center}
    \textit{Supplementary Material}
\end{center}
\url{https://github.com/MoBaioumy/active_inference_panda_paper}.


\section{INTRODUCTION}
\label{sec:introduction}
It is crucial for intelligent robots to be able to adapt to the presence of unmodeled dynamics. This is necessary for applications such as aerial vehicles encountering unpredicted wind dynamics \cite{sierra2019wind}, manipulators handling objects of unknown weight \cite{wei2018adaptive}, and autonomous vehicles on slippery road surfaces \cite{arifin2019lateral}. Humans are skilled in this regard and recent work in robotics has taken inspiration from \textit{active inference} \cite{friston2017active}, a  neuroscientific theory of the brain and behaviour. Active inference provides a framework for understanding decision-making of biological agents where optimal behavior arises from minimising variational free-energy: a measure of the fit between an internal model and (past) sensory observations. Additionally, agents act in a fashion that fulfills prior beliefs about preferred future observations \cite{FEProughbrainguide}. This framework has been employed to explain and simulate a wide range of complex behaviors including abstract rule learning, speech generation, planning and navigation \cite{pezzulo2018hierarchical,kaplan2018planning,friston2017active}.

\begin{figure}
    \centering
    \includegraphics[width=\linewidth]{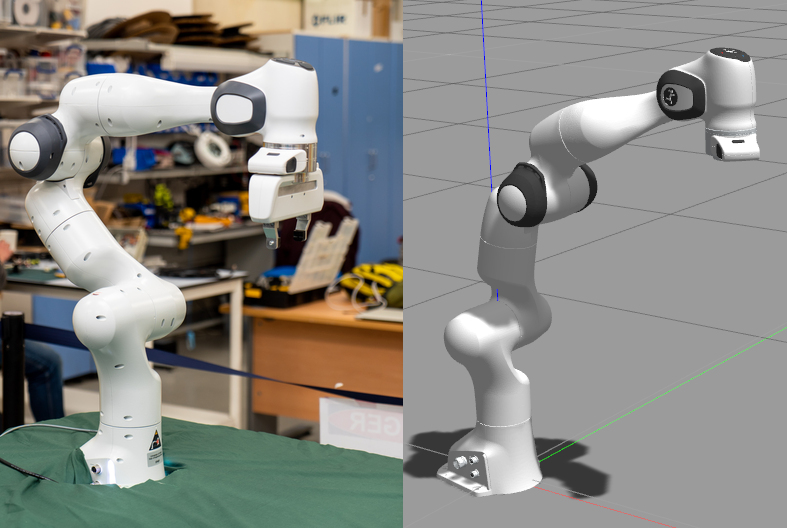}
    \caption{The Franka Emika  Panda  7  DoF  manipulator real robot (left) and simulated robot in Gazebo (right).}
    \label{fig:pandarobot}
\end{figure}

The active inference controller (AIC) \cite{buckley2017free}, allows an agent to jointly perform state-estimation and control by minimizing a single quantity: variational free-energy. Subsequent work has utilized the AIC in robotics \cite{lanillos2018active, oliver2019active, pezzato2019novel}. The control and state-estimation parts are \textit{coupled} and the AIC architecture is unusual compared to methods commonly used in robotics. Usually, when performing state-estimation (for instance using a Kalman or particle filter \cite{kalman1960new, gordon1993novel}) observations are required. The aim is then to find a belief state that is close to the true (hidden) state. Additionally, controllers (such as a PID \cite{aastrom1995pid}) require a target state a withlong the current state. A PID controller then provides a control law to reach the target from the current state. In the AIC however, state-estimation requires the target and the observations which is usual from a control perspective; those are used to produces a `biased' belief. The control part of the AIC requires the observation and the `biased' belief. It is thus unclear how to systematically tune such a controller (since tuning the controller affects state-estimation). Furthermore, even slight changes to the parameters of the state-estimation component could cause the controller to be unstable.

This paper has three contributions. First, we present a formulation which includes a temporal parameter $\tau$. Second, we highlight theoretical results showing the exact relationship between our formulation and existing methods. For example, if our introduced parameter $\tau$ approaches zero, the approach converts to a classic PID controller. When $\tau \rightarrow \infty$, the approach converts to a filter. If $\tau$ is set to the unity, the formulation presented in \cite{buckley2017free} is recovered. The third contribution is presenting a method to automatically tune the controller. We show that the gains of the controller correspond to the covariances of the observation model and that learning the optimal covarinaces results in learning the optimal gains for the controller. However, the covariances do not converge when the systems reaches the target. We thus present an alternative by learning the introduced temporal parameter $\tau$. This reduces oscillations and improves robustness. 

As a running example throughout the paper, the  mass-spring-damper~\cite{dahleh1998theory} system is used. However, we validate the approach on a `Franka Emika Panda' 7 DoF manipulator in the results section.



\section{Related work}
\label{related_work}
There are only a few attempts to use active inference in robotics and control. In \cite{pio2016active}  a  simulated PR2  robot is controlled  by  open-loop  active inference but the computational complexity made an online  implementation unfeasible.  In \cite{lanillos2018adaptive}, the authors use free-energy minimization for adaptive body perception. The same authors extended their work to include control in \cite{lanillos2018active}. In \cite{oliver2019active} an  implementation of active inference is  presented  with  real  hardware on a humanoid robot capable of performing reaching  behaviors  with  both  arms  along with  active  head  object tracking in the presence noisy observations. This work is extended in \cite{lanillos_sancaktar2019end} using deep neural networks as function approximators.

In \cite{pezzato2019novel} an approach for control of  robotic manipulators is presented. Adaptive behaviour is demonstrated against the state-of-the-art model-reference adaptive control (MRAC)~\footnote{Model-reference adaptive control \cite{zhang2017review},  finds a control law that will guide the system to behave as specified by a chosen reference model. Another common method used for adaptive control is `self-tuning adaptive control',  which represent the robot as a linear discrete-time model and estimates the unknown parameters online, substituting them in the control law \cite{walters1982application}.}. 
Additionally, in \cite{2020IWAI_baioumy}, the authors use active inference to achieve fault-tolerant control for sensory faults \cite{ZHANG2008229,surveyFT}. The robot can detect whether its sensors are faulty and the control law is automatically adjusted to account for the detected faults. In \cite{anil_colored_noise}, free-energy minimization is used to improve state-estimation under coloured noise. In \cite{SimulatingVMP, vander_laar_2019_lqg, vanderbroeck2019active} active inference with factor graphs is proposed as an effective control method; however, results are provided only for a simulated toy example.


\section{Active Inference framework}
\label{Actinve inference and free energy}
This section introduces active inference as a general framework and derives the key equations for free-energy ($F$). Free-energy is used in later sections to achieve state-estimation, control and hyperparameter learning.

\subsection{Variational free energy}

We consider an agent in a dynamic environment that receives observation $\Observation$ about the hidden state $\state$. Given a model of the agent's world, Bayes' rule can be used to find $p(\state|\Observation)$. 
However, the normalization term $p(\Observation)$ in Bayes' rule involves calculating an integral making calculations of all but trivial examples infeasible. Instead, the agent can approximate the posterior distribution $p(\state|\Observation)$ with a \textit{variational distribution} $\variationaldistribution$ over states, which we can define to have a simpler form (such as a Gaussian). The goal is then to minimize the difference between the two distributions. The mismatch between the two distribution can be computed using the Kullback Leibler (KL) divergence \cite{fox2012tutorial, wainwright2008graphical}: 

\begin{equation}
\begin{split}
    \label{eq:Kullback-leibler}
    KL(\variationaldistribution(\state)|| p(\state|\Observation)) &= \int{\variationaldistribution(\state) \ln{\frac{\variationaldistribution(\state)}{p(\state, \Observation)}}
    d\state + \ln p(\Observation)}\\
    &= F + \ln p(\Observation).
\end{split}
\end{equation}

The quantity $F$ is referred to as the (variational) free-energy and minimizing F minimizes the KL-divergence. $F$ is also often referred to as the evidence lower bound (ELBO) in the Machine Learning community\footnote{Minimizing the ELBO and thus the KL divergence is common in variational inference, a method for approximating probability densities \cite{fox2012tutorial}.}. If we choose $\variationaldistribution$($\state$) to be a Gaussian distribution with mean $\mu$, and utilize the Laplace approximation \cite{bishop2006pattern}, the free-energy expression from Equation \ref{eq:Kullback-leibler} simplifies to:
\begin{equation}
\begin{split}
    \label{eq:laplace_F}
    F \approx &  -\ln p(\mu, \Observation). 
\end{split}
\end{equation}

The expression for $F$ is solely dependent on one parameter, $\mu$, which is referred to as the `belief state' or simply `belief'. The objective is to find the $\mu$ which minimizes $F$ and thus minimizing the divergence between $q(s)$ and $p(s|o)$. For a robotic manipulator the set of observations ($\mathbf{\Observation}$) and beliefs ($\bm{\mu}$) are vectors with length depending on the number of degrees of freedom.


Generalised motions (GM) \cite{friston2008hierarchical} are used to represent the (belief) states of a dynamical system, using increasingly higher order derivatives of the system state. This means that the n-dimensional state $\bm{\mu}$ and its higher order time derivatives are combined in $\bm{\tilde{{\mu}}}$  (i.e. $\bm{\tilde{\mu}} = [\bm{\mu}, \bm{\mu'} ...]$). Similarly, observations are combined as $\mathbf{\tilde{\Observation}} = [\mathbf{\Observation}, \mathbf{\Observation'}$ ...]. In the context of a robotic manipulator, $\mathbf{\Observation}$ may represent the sensory observation of joint positions, while $\mathbf{\Observation'}$ represents the observation of joint velocities. Similarly, $\bm{\mu}$ represents the belief about all joint positions and $\bm{\mu'}$ the joint velocities.

\subsection{Observation model and state transition model}
\label{sec:models_gen_and_sensory}

Taking generalized motions into account, the joint probability from Equation~(\ref{eq:laplace_F}) can be written as:

\begin{equation}
\begin{split}
    \label{eq:generative_model_p(mu,O)}
    p(\tilde{\mathbf{\Observation}}, \bm{\tilde{\mu}})
    &= p(\tilde{\mathbf{\Observation}}| \bm{\tilde{\mu}})p(\bm{\tilde{\mu}}) =p(\mathbf{\Observation}| \bm{\mu})p(\mathbf{\Observation'}| \bm{\mu'})p(\bm{\mu'}| \bm{\mu})p(\bm{\mu''}| \bm{\mu'}),
\end{split}
\end{equation}
where $p(\mathbf{\Observation}| \bm{\mu})$ is the probability of receiving an observation $\mathbf{\Observation}$ while in (belief) state $\bm \mu$, and $p(\bm{\mu'}| \bm{\mu})$ is the state transition model (also referred to as the dynamic model or the generative model). The state transition model predicts the state evolution given the current state. These distributions are assumed Gaussian according to:
\begin{equation}
\begin{split}
    \label{eq:gaussian_model_for_gen_and_sen}
    p(\mathbf{\Observation}| \bm{\mu}) = \mathcal{N}(g(\bm{\mu}), \Sigma_\Observation), & 
    \;\; p(\mathbf{\Observation'}| \bm{\mu'}) = \mathcal{N}(g'(\bm{\mu'}), \Sigma_{\Observation'}), \\
    p(\bm{\mu'}| \bm{\mu}) = \mathcal{N}(f(\bm{\mu}), \Sigma_{\mu}), &
   \;\; p(\bm{\mu''}| \bm{\mu'}) = \mathcal{N}(f'(\bm{\mu'}), \Sigma_{\mu'}),
\end{split}
\end{equation}
where the functions $g(\bm{\mu})$ and $g'(\bm{\mu'})$ represent a mapping between observations and states. For many applications in robotics the state is directly observable. For instance, in the context of a robotic manipulator the state consists of the positions and velocities of all joints and the manipulator is provided with position and velocity encoders. Thus we assume:  $g(\bm{\mu}) = \bm{\mu}$ and $g'(\bm{\mu'}) = \bm{\mu'}$. 
The functions $f(\bm{\mu})$ and $f(\bm{\mu'})$ represent the evolution of the belief state over time. We encode the agent's target state $\bm{\mu_{d}}$ in $f(\bm{\mu})$. Our contribution is to introduce $\tau$ as part of these funcitons: $f(\bm{\mu}) = (\bm{\mu_{d}} - \bm{\mu})\tau^{-1}$ and $f'(\bm{\mu'}) = \tau^{-1}\bm{\mu'}$, where $\bm{\mu_{d}}$ is the desired state and $\tau$ a time scale (explained in Section \ref{Understanding temporal parameter}).

To simplify notion, we define the following error terms: $\bm{\varepsilon_{\mu}} = \bm{\mu'} - (\bm{\mu_{d}} - \bm{\mu})\tau^{-1}$, 
$\bm{\varepsilon_{\mu'}} = \bm{\mu''} + \tau^{-1}\bm{\mu'}$, 
$\bm{\varepsilon_{\Observation}} = \mathbf{\Observation} - \bm{\mu}$,  
$\bm{\varepsilon_{\Observation'}} = \mathbf{\Observation'} - \bm{\mu'}$. Now that all the terms have been defined, $F$ can be expanded to:
\begin{align}
        \label{eq:laplace_F_final_vector}
    	F = \frac{1}{2}\sum_i\begin{pmatrix} \bm\varepsilon_i^\top \Sigma_{i}^{-1} \bm\varepsilon_i + \ln{|\Sigma_{i}|}\end{pmatrix} + C,
\end{align}
\noindent where $i\in~\{\Observation,\ \Observation',\ \mu,\ \mu'\}$, $C$ is a constant, and the covariance matrices are defined in Equation~\ref{eq:gaussian_model_for_gen_and_sen}.  Equation~(\ref{eq:laplace_F_final_vector}) differs from the work presented in \cite{pezzato2019novel, oliver2019active} in the terms with $\ln|\Sigma|$, which are not explicitly included but rather added to the constant. The significance of this difference is highlighted in Section \ref{Learning model variances}.

\section{State-estimation and Control}
\label{Control and state-estimation}
We now introduce how to perform state-estimation and control by minimizing $F$. We show how the estimation step biases the belief towards the goal state. The control step then steers the system from its observation $\mathbf{\Observation}$ to its (biased) belief~$\bm{\mu}$. In Section \ref{subsec:relationship to PID} we show that if $\tau^{-1} \rightarrow \infty$, the approach converts to a classic PID controller. Additionally, if $\tau^{-1} \rightarrow 0$, the approach reduces to a filter.

\subsection{State estimation by minimizing free-energy}

Estimating the state of our system is achieved by finding a value $\bm \mu$ that minimizes $F$. Gradient descent is a simple way to accomplish that:

\begin{equation}
    \label{eq:F_minimize_mu}
    \bm{\dot{\tilde{\mu}}} = D\bm{{\tilde{\mu}}} - \kappa_{\mu}\frac{\partial F}{\partial {\bm{\tilde{\mu}}}},
\end{equation}

\noindent where $\kappa_{\mu}$ is a tuning parameter and $D$ is temporal derivative operator ($D \bm \mu = \bm \mu'$). The dot refers to the difference between two time-steps ($\dot{\bm{\mu}} = \bm{\mu}[t+1] - \bm{\mu}[t]$). Using Equation (\ref{eq:F_minimize_mu}) the agent takes one-step in the gradient descent at every time-step. In this case the equation expands to: 

\begin{equation}
\begin{split}
    \label{eq:mu_update_rules_mu_vector}
    \dot{\bm{\mu}} &= \bm{\mu'} + \kappa_{\mu} \Sigma_{\Observation}^{-1} \bm{\varepsilon_{\Observation}} - \tau^{-1}  \kappa_{\mu}\Sigma_{\mu}^{-1}\bm{\varepsilon_{\mu}} \\
    \dot{\bm{\mu'}} &= \bm{\mu''} + 
    \kappa_{\mu} \Sigma_{\Observation'}^{-1} \bm{\varepsilon_{\Observation'}}
    - \kappa_{\mu}\Sigma_{\mu}^{-1}\bm{\varepsilon_{\mu}} 
    - \tau^{-1}\kappa_{\mu'}\Sigma_{\mu'}^{-1}\bm{\varepsilon_{\mu'}}\\
    \dot{\bm{\mu''}} &= - \kappa_{\mu'}\Sigma_{\mu}^{-1}\bm{\varepsilon_{\Observation'}}
\end{split}
\end{equation}

The first equation states that belief is refined using the term $\kappa_{\mu} \Sigma_{\Observation}^{-1} \bm{\varepsilon_{\Observation}}$ which moves our new belief towards the value just observed. Additionally, the term $\tau^{-1}  \kappa_{\mu}\Sigma_{\mu}^{-1}\bm{\varepsilon_{\mu}}$, shifts the belief towards the target $\bm{\mu_d}$ since $\bm{\varepsilon_{\mu}} = \bm{\mu'} - (\bm{\mu_{d}} - \bm{\mu})\tau^{-1}$. Essentially, this `biases' the belief towards preferred future states (target state $\bm{\mu_{d}}$). The degree to which the system is biased depends on the the values $\tau^{-1}$ and $\Sigma_{\mu}^{-1}$.

\subsection{Control by minimizing free-energy}
\label{subsec:Control by minimizing free-energy}

Next to state-estimation, the agent can apply an action $\mathbb{a}$ to the environment. In the context of a robotics manipulator this is a vector containing torques of all joints. To find the control action which minimizes $F$, gradient descent is used: 
\begin{equation}
    \label{eq:F_minimize_a}
    \mathbf{\dot{\action}} = - \kappa_{\action}\frac{\partial F}{\partial \mathbf{\action}}
    = - \kappa_{\action} \frac{\partial F}{\partial \tilde{\mathbf{\Observation}}}  \frac{\partial \tilde{\mathbf{\Observation}}}{\partial \mathbf{\action}},
\end{equation}

\noindent where $\kappa_{\action}$ is a tuning parameter. The term $\frac{\partial \tilde{\mathbf{\Observation}}}{\partial \mathbf{\action}}$ is assumed linear, and equal to the identity matrix (multiplied by a constant) similar to existing work ~\cite{pezzato2019novel, oliver2019active}. Actions are then computed as:
\begin{equation}
    \label{eq:update_rules_a}
    \mathbf{\dot{\action}} = -\kappa_{\action}(\Sigma_{\Observation}^{-1} (\mathbf{\Observation} - \bm{\mu}) + \Sigma_{\Observation'}^{-1} (\mathbf{\Observation'} - \bm{\mu'})).
\end{equation}

This controller essentially steers the system from its observed state $\mathbf{\Observation}$ to its (biased) belief $\bm{\mu}$.

Note how the current control law does not contain any information about the dynamical system, it is thus a reactive controller. The control law only requires $\mathbf{\Observation}$ and $\bm{\mu}$. This controller thus operates in the presence of unmodeled dynamics similar to a PID controller.

\subsection{Simultaneous state-estimation and control}
Our approach performs state estimation and control simultaneously. The estimation and control step are dependent. This is because the estimation step biases the belief $\bm{\mu}$ towards the target $\bm{\mu_d}$. The controller then steers the system from the observation $\mathbf{\Observation}$ to the biased estimated state $\bm{\mu}$. If $\tau^{-1}$ and $\Sigma_{\mu}^{-1}$ are large, the belief $\bm{\mu}$ is biased more towards the target $\bm{\mu_d}$.

\begin{figure}
    \centering
    \includegraphics[width=\linewidth]{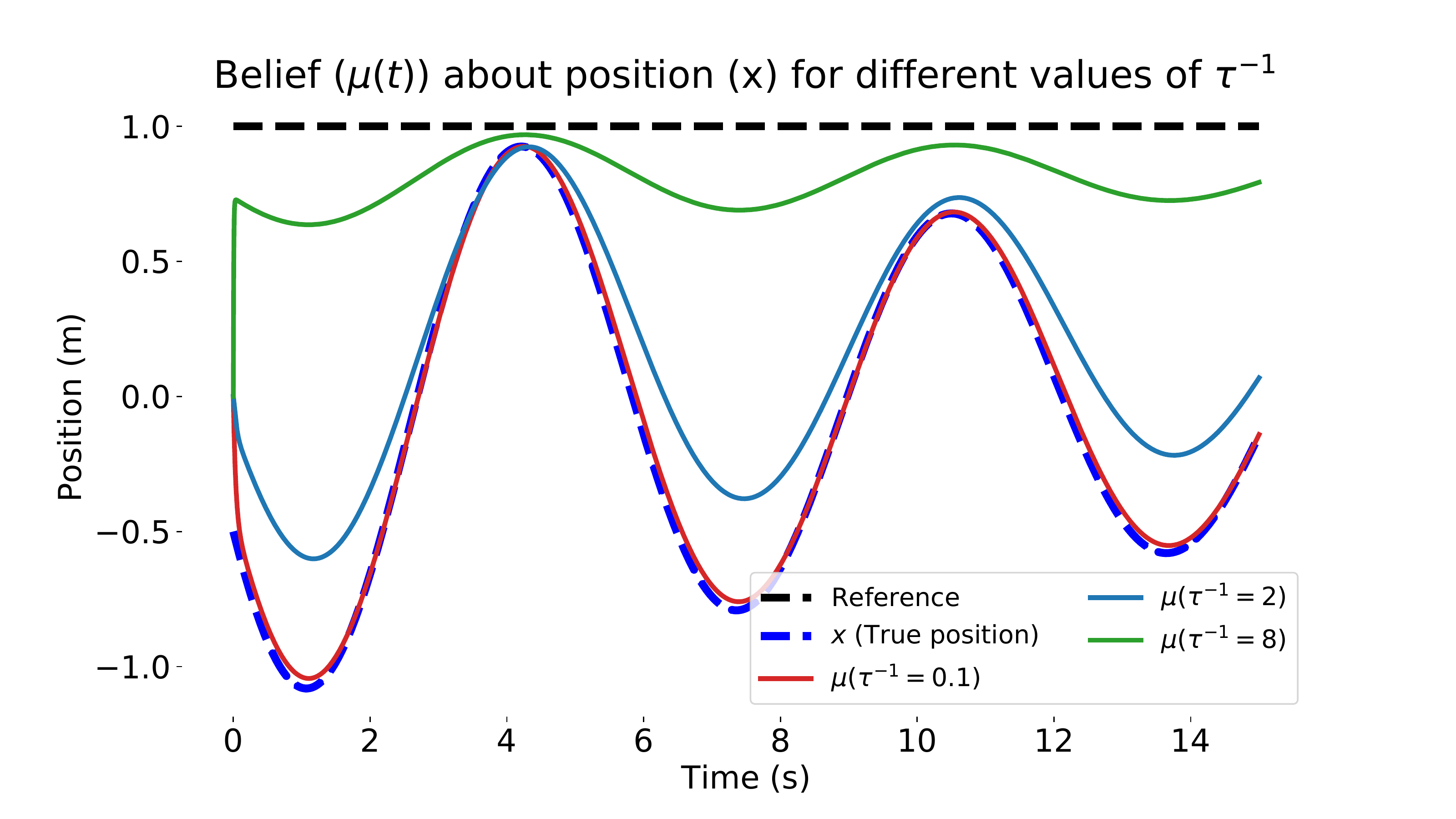}
    \caption{Separately performing state-estimation for different values of $\tau^{-1}$. Higher values of $\tau^{-1}$ result in more bias.}
    \label{fig:estimation_only_differnt_delta_values}
\end{figure}

To illustrate this, consider the mass-spring-damper system given by the equation: $\ddot{x} = a(t) - k_1x - k_2\dot{x}$, where $s$ is the position of the mass, $a(t)$ the control action, $k_1$ the spring constant (set to $1 N/m$), $k_2$ the damping coefficient (set to $0.1 Ns/m$) and the system has unit mass. It's simulated with initial conditions $x(0) = -0.5m$, $\dot{s}(0) = -1 m/s$ and $a(t) = 0N$. Equations (\ref{eq:mu_update_rules_mu_vector}) are used to perform state estimation. To challenge the system, the initial beliefs are inaccurate ($\bm{\mu}(0) = 0m$ and $\bm{\mu'}(0) = -1.5m/s$). The system is simulated for different values of $\tau^{-1}$ and presented in Figure \ref{fig:estimation_only_differnt_delta_values}. 


It is clear that higher values of $\tau^{-1}$ provide more bias towards the target. For $\tau^{-1} = 8$ (green line in Figure \ref{fig:estimation_only_differnt_delta_values}), the estimate is close to the target (black dashed line) and far away form the actual position (blue dashed line) as opposed to setting $\tau^{-1} = 0.1$ (red line), the belief is closer to the real trajectory (the latent state). If $\tau^{-1} \rightarrow  0$, the estimation step reduces to a pure estimator, which would follow the trajectory without any bias towards the target.

\begin{figure}
    \centering
    \includegraphics[width=\linewidth]{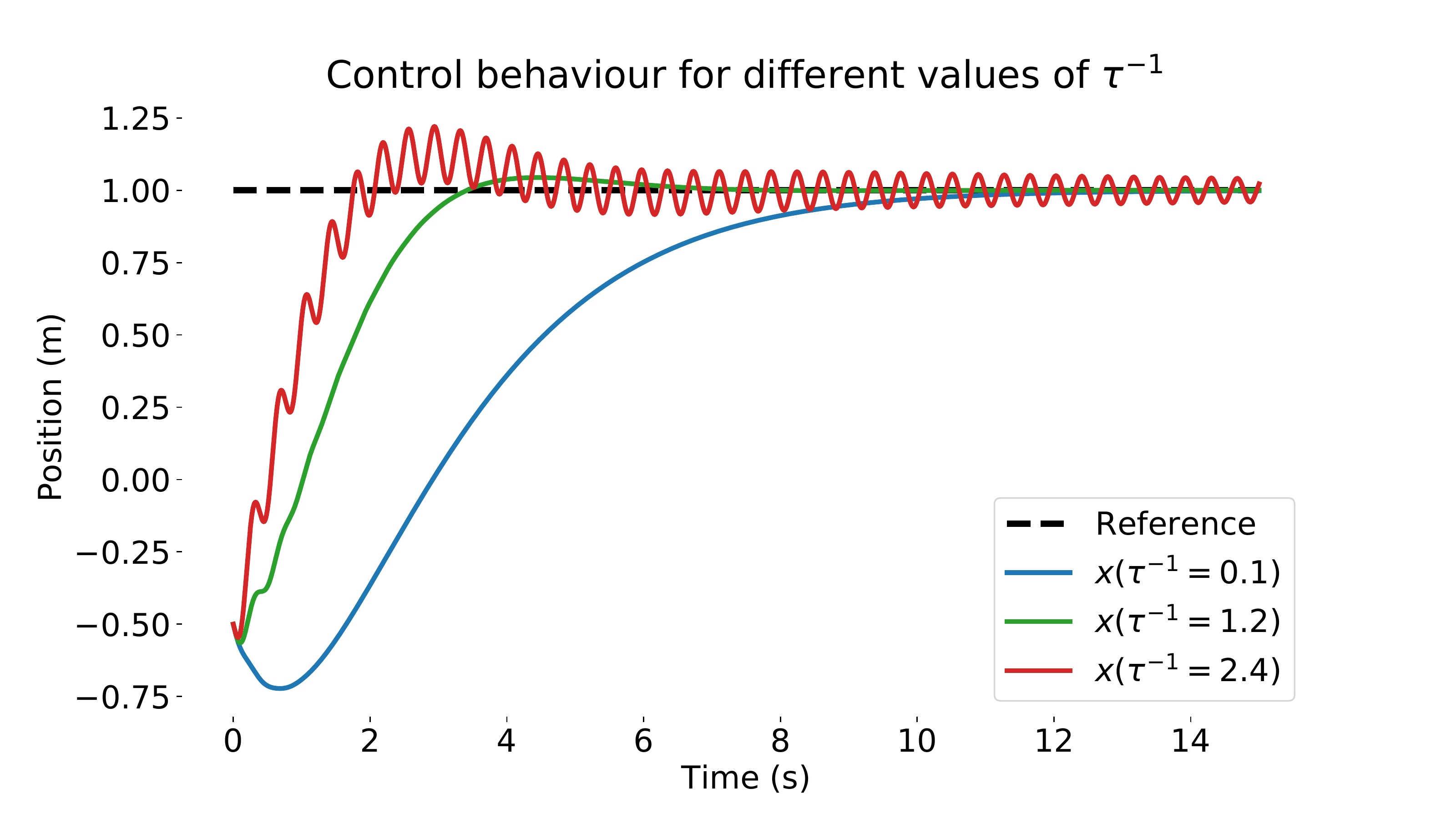}
    \caption{True position ($x$) for different values of $\tau^{-1}$. Higher values of $\tau^{-1}$ provide more bias towards the target and thus more aggressive control causing overshoot oscillations.}
    \label{fig:control_different_delta_values}
\end{figure}

Enabling control steers the system to its target. The $\tau^{-1}$ in this case affects how aggressive the controller is. Larger values give more bias towards the target, the term $(\mathbf{\Observation} - \bm{\mu})$ is larger and thus the controller is more aggressive. Figure \ref{fig:control_different_delta_values} shows an illustration for different values of $\tau^{-1}$.

\subsection{Understanding the temporal parameter $\tau$}
\label{Understanding temporal parameter}
The generative model specified by Equations (\ref{eq:generative_model_p(mu,O)}) and (\ref{eq:gaussian_model_for_gen_and_sen}) include the function $f(\bm{\mu})$ which determines how the belief state evolves over time i.e.  $f(\bm{\mu}) = (\bm{\mu_{d}} - \bm{\mu})\tau^{-1}$. 

The belief state is specified to evolve over time as the derivative between the current belief $\bm{\mu}$ and target $\bm{\mu_d}$. This can be evaluated as the ($\bm{\mu_d}$ - $\bm{\mu}$) divided by a time scale $\tau$. The smaller $\tau$, the larger the derivative. If $\tau$ approaches zero ($\tau^{-1} \rightarrow \infty$), the value $f(\bm{\mu})$ approaches $\infty$. As a results, the belief is infinitely biased towards the target and $\bm{\mu}\approx\bm{\mu_d}$.


\subsection{Relationship to a classic PID Controller and filters}
\label{subsec:relationship to PID}
A classic PID controller defines an error term $e = (\bm{\mu_d} - \mathbf{\Observation})$. The control action is then chosen as: $$ a = P\cdot e + I\int{e} dt+D\frac{de} {dt}, $$ where $P$,  $I$ and $D$ are tuning parameters. 

For the control law defined by active inference, our $(\mathbf{\Observation} - \bm{\mu})$ is similar to the error term. Additionally, as explained in the previous section, when $\tau^{-1} \rightarrow  \infty$ then $\bm{\mu}\approx\bm{\mu_d}$. Now the control law of active inference can be rewritten in terms of the error term as:
$$ \dot{a} = \kappa_{\action} \Sigma_{\Observation}^{-1} e  + \kappa_{\action} \Sigma_{\Observation'}^{-1} \frac{de}{dt}.$$

This means than if $\tau^{-1} \rightarrow \infty$, the active inference controller is equivalent to a PI Controller i.e. PID with $D=0$, a $P$ gain of $\kappa_{\action} \Sigma_{\Observation'}^{-1}$ and an $I$ gain of $\kappa_{\action} \Sigma_{\Observation}^{-1}$. If one considers the generalized motions (section \ref{Actinve inference and free energy}) up to a third order, the control law would include a non-zero $D$ term.

The relationship to a pure estimator (plain filter) is straightforward. As previously mentioned, if $\tau^{-1} \rightarrow  0$, the estimation step reduces to a filter. Essentially, this indicates that the estimation step has zero bias towards the target. As Figure \ref{fig:estimation_only_differnt_delta_values} shows that for very small values of $\tau^{-1}$, the belief is close to the latent state (true position) without bias.

\section{Learning the hyperparameters}
\label{sec:learning_as_active_inference}
We have shown that state estimation and control can be performed using gradient decent on the free-energy $F$. In addition to using gradient decent of the free energy for state-estimation and control, we apply it to learn the hyperparameters online. Previous work such as \cite{pezzato2019novel} does not update the hyperprameters online which make the performance extremely sensitive to initialization, prone to instabilities and reduces overall performance. 

\subsection{Learning model variances}
\label{Learning model variances}
As illustrated in Section \ref{subsec:relationship to PID}, the model variances $\Sigma_{\Observation}^{-1}$ and $\Sigma_{\Observation'}^{-1}$ can be considered as gains for the controller, similar to the `P' and `I' gains in a PID controller. Additionally, the values $\Sigma_{\mu}^{-1}$ and $\Sigma_{\mu'}^{-1}$ affect how much the estimation step biases the controller towards the desired position. 

We can update $\Sigma_{\Observation}$ and $\Sigma_{\Observation'}$ using gradient decent on $F$ as:

\begin{equation}
    \label{eq:observaion_variance_update_without_inverse}
    \dot{\Sigma}_{\Observation} = -\kappa_{\sigma} \frac{\partial F}{\partial \Sigma_{\Observation}}, \hspace{3mm} \dot{\Sigma}_{\Observation'} = -\kappa_{\sigma} \frac{\partial F}{\partial \Sigma_{\Observation'}}.
\end{equation}

The presented update rules have several practical issues. First, in any high dimensional case, $\Sigma_{\Observation}$ would be a matrix. Since in most equations presented so far, the inverse $\Sigma_{\Observation}^{-1}$ is used, updating the covariance matrix using Equations \ref{eq:observaion_variance_update_without_inverse} then inverting it would be computationally expensive. A workaround is to simply update the inverse covariance matrix (the precision matrix), as done in \cite{2020UKRAS_baioumy}: 

\begin{equation}
    \label{eq:observaion_variance_update}
    \dot{\Sigma}_{\Observation}^{-1} = -\kappa_{\sigma} \frac{\partial F}{\partial \Sigma_{\Observation}^{-1}}, \hspace{3mm} \dot{\Sigma}_{\Observation'}^{-1} = -\kappa_{\sigma} \frac{\partial F}{\partial \Sigma_{\Observation'}^{-1}}.
\end{equation}

Additionally, a lower-bound on the diagonal elements is set to keep the matrix positive semi-definite as suggested in \cite{bogacz2017tutorial}.

\begin{figure}
    \centering
    \includegraphics[width=0.95\linewidth]{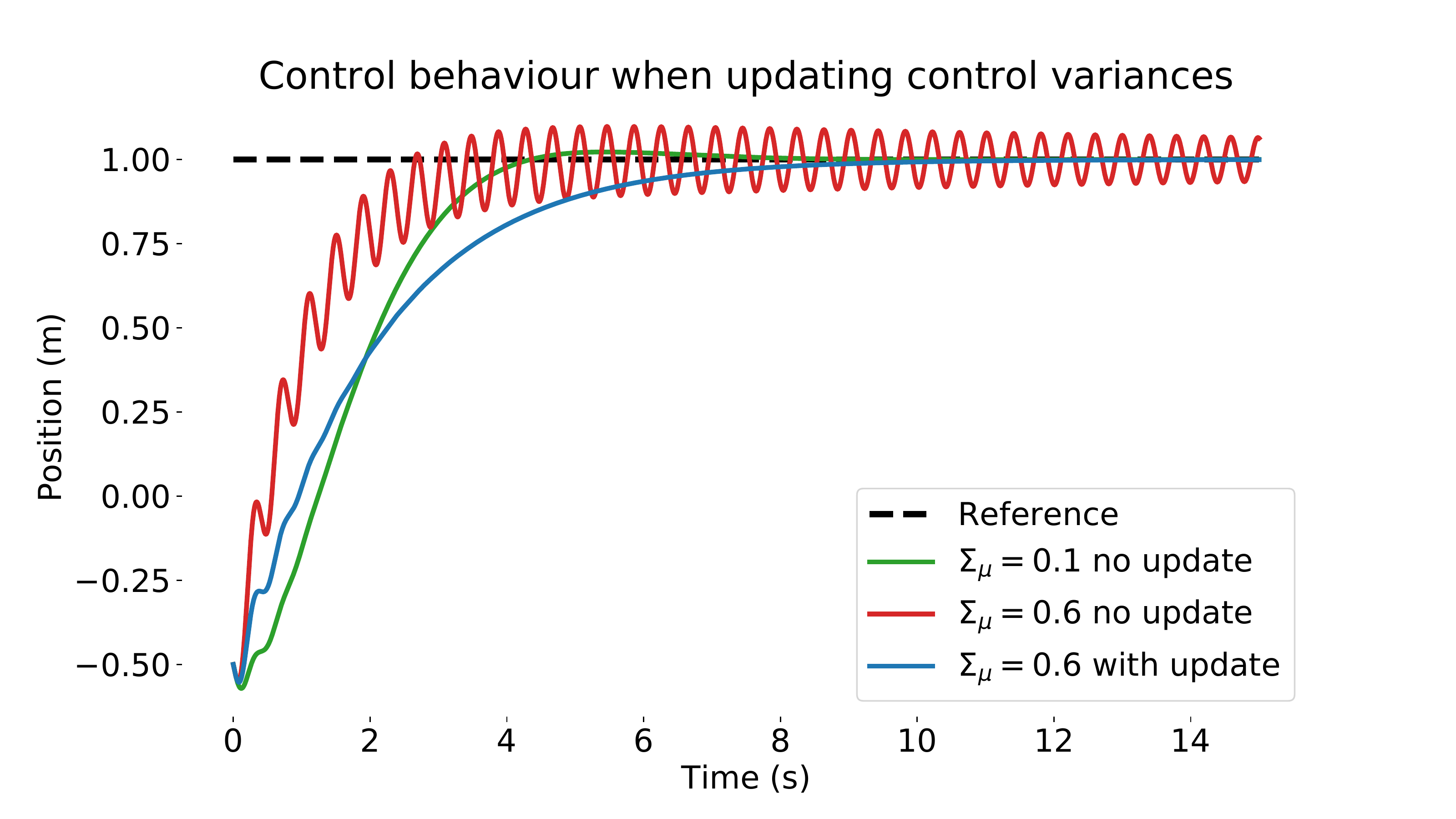}
    \caption{Effect of updating the the control variances ($\Sigma_{\Observation}^{-1}$ and $\Sigma_{\Observation'}^{-1}$). When the value of $\Sigma_{\mu}^{-1}$ is initialized at high values, the system oscillates. Updating $\Sigma_{\Observation}^{-1}$ and $\Sigma_{\Observation'}^{-1}$ essentially `tunes' the controller and ensures robust performance.}
    \label{fig:control_updating_observation_variance}
\end{figure}

We demonstrate the effect of updating the covarince by keeping $\Sigma_{\mu'}^{-1}$ fixed at 0.5 and $\Sigma_{\mu}^{-1}$ will be varied. If  $\Sigma_{\mu}^{-1}$ is too high, the systems suffers from oscillations and overshoot. However, if $\Sigma_{\Observation}^{-1}$ and $\Sigma_{\Observation'}^{-1}$ are updated during run-time, the controller shows improved behaviour. Results are shown in Figure \ref{fig:control_updating_observation_variance}. The convergence of $\Sigma_{\Observation}^{-1}$ occurs when $\frac{\partial F}{\partial \Sigma_{\Observation}^{-1}} = 0$. Since the observations change and have a certain level of noise, $\Sigma_{\Observation}$ converges to the expected value of $\bm{\varepsilon_{\Observation}}\bm{\varepsilon_{\Observation}}^\top$. This does not necessarily happen upon reaching the target state.






\begin{figure}
    \centering
    \includegraphics[width=0.95\linewidth]{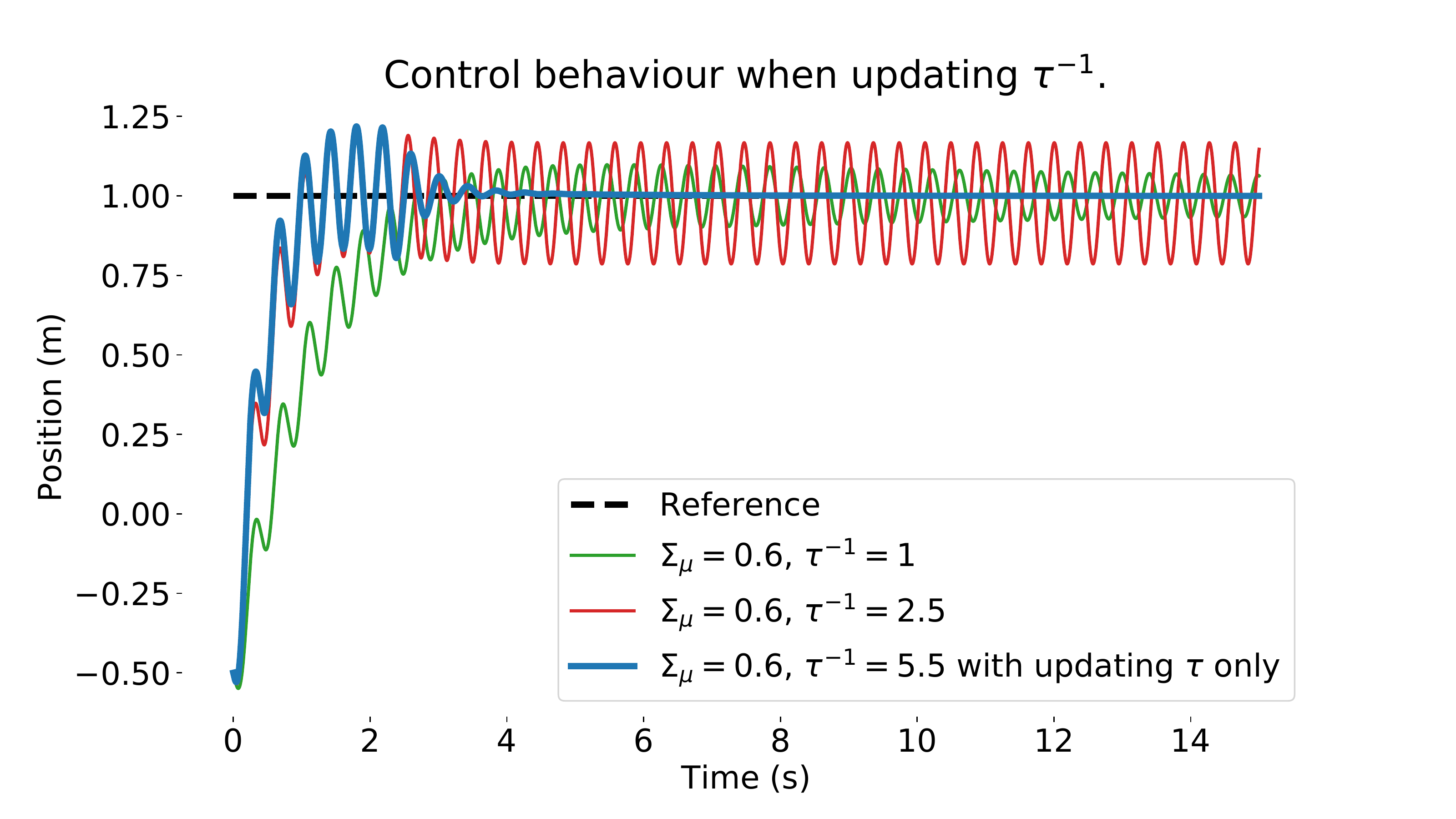}
    \caption{Effect of updating the value of $\tau^{-1}$ during operation. This figure shows that increasing both $\Sigma_{\mu}$ or $\tau^{-1}$ results in overshoots and severe oscillations. Rather than tuning  $\Sigma_{\Observation}^{-1}$ and $\Sigma_{\Observation'}^{-1}$, updating $\tau^{-1}$ can be sufficient. }
    \label{fig:control_updating_delta}
\end{figure}

\begin{figure*}[!t]
  \includegraphics[width=0.95\textwidth]{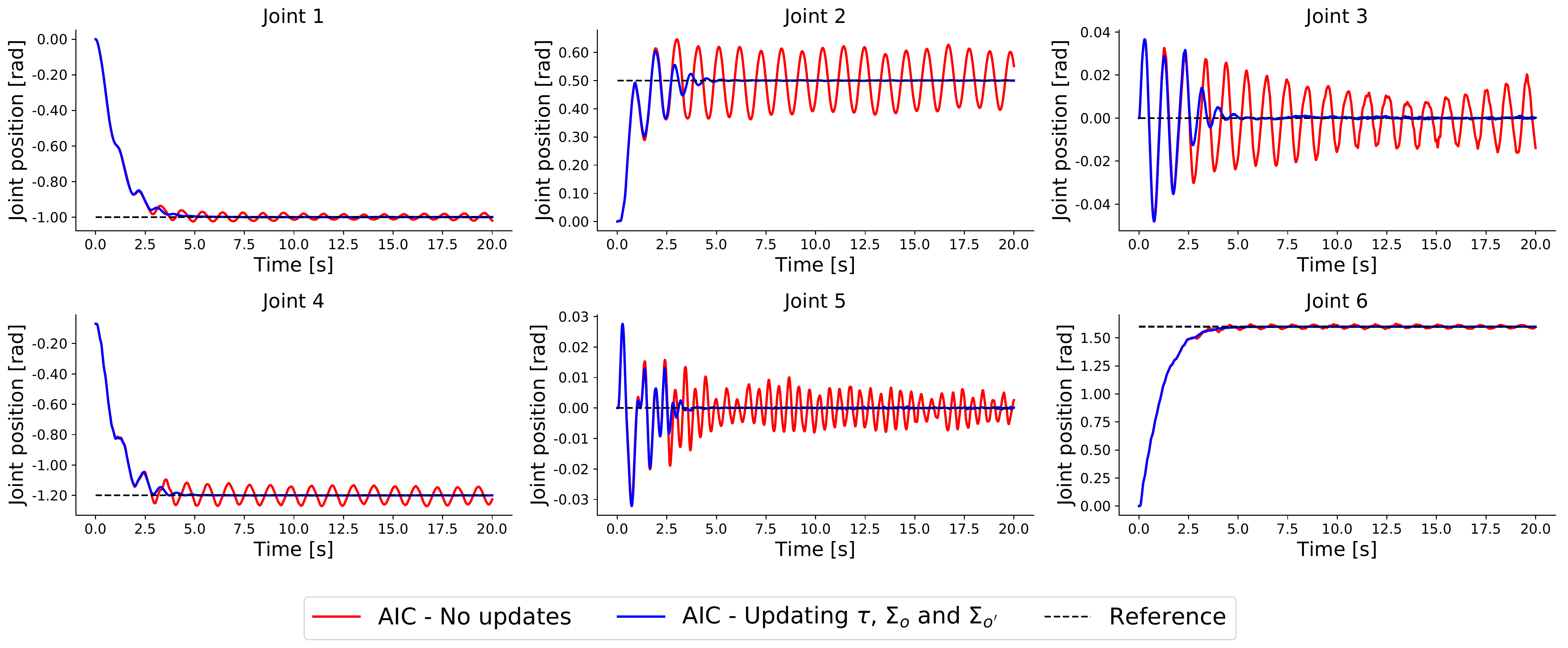}
  \caption{Step response for the 6 joints of the robot arm  with and without updating hyperparameters in the AIC. This graphs corresponds to the second column of Table \ref{table:varying simga} ($\Sigma_{\mu}^{-1} = 0.3I$). It is clear the updating the hyperparameters reduces oscillations.}
  \label{fig:results for the real pands arm}
\end{figure*}


\subsection{Learning the temporal parameter}
We previously showed the importance importance of choosing appropriate values for $\tau^{-1}$: If the value is too high, the controller suffers from overshoot and oscillations, see Figure \ref{fig:control_different_delta_values}. On the other hand, a low value results in a slow response. Ideally, the value for $\tau$ would be high towards the start but decrease as the system reaches the target. This is the value that minimizes $F$ and can be found using gradient descent on $F$ as: 

\begin{equation}
\begin{split}
    \label{eq:minimize tau}
    \frac{\partial F}{\partial \tau^{-1}} &= -2 \Sigma_{\mu'}\varepsilon_{\mu} (\bm{\mu_d} - \bm{\mu}) + 2\Sigma_{\mu'}\varepsilon_{\mu'}\bm{\mu'}.
\end{split}
\end{equation}

\noindent Note how the inverse $\tau^{-1}$ is updated rather than $\tau$ directly. Similar to the variances, all previous equations contain $\tau^{-1}$ and since inverting a matrix is computationally expensive, the inverse is directly updated. Additionally, $\tau^{-1}$ requires us to define a lowerbound. The optimization can results in $\tau^{-1}$ approaching zero
which means the controller converts to a pure estimator. In this work, $\tau^{-1}$ is set to have a minimum value of 0.5 on the diagonal elements and zero elsewhere.

Using Equation \ref{eq:minimize tau}, the oscillations can be damped as well as improving settling time as shown in Figure \ref{fig:control_updating_delta}. Note how updating $\tau^{-1}$  only is satisfactory to eliminate the oscillations (no update of $\Sigma_{\Observation}^{-1}$ or $\Sigma_{\Observation'}^{-1}$ is necessary).
The convergence of $\tau^{-1}$ occurs  at $\bm{\mu_{d}} = \bm{\mu}$ and $\bm{\mu'} = 0$ which corresponds to the controller settling at its target position. The updates for $\tau$ will thus retune the controller appropriately until the the target is reached. This provides a preference for updating $\tau$ rather than updating $\Sigma_{\Observation}^{-1}$ or $\Sigma_{\Observation'}^{-1}$ in most cases.

\subsection{Limitations}
In the active inference setting, the belief is intentionally biased towards the target to achieve control. However, the belief is biased and thus not accurate unless the system reaches the target. If an accurate belief is required for other parts of the robotic system, this could be problematic. 

Additionally, the terms $\Sigma_{\Observation}^{-1}$ and $\Sigma_{\Observation'}^{-1}$ are defined as part of the sensory model. In the context of a filter, this would indicate the level of Gaussian noise affecting the position and velocity encoders. However, since the beliefs are biased, $\Sigma_{\Observation}^{-1}$ and $\Sigma_{\Observation'}^{-1}$ converge to a value that is much higher than the actual Gaussian noise affecting the system. The values for these variables therefore lose their physical meaning. This bias can also cause false-positives when reasoning about the accuracy of the sensor. Work in \cite{2020IWAI_baioumy} explores this problem. 

Finally, since estimation and control are dependent on each other, it is not straight-forward to tune all the parameters separately.  Learning the parameters $\Sigma_{\Observation}^{-1}$ and $\Sigma_{\Observation'}^{-1}$ boosts the performance as shown in the previous section. However, if $\Sigma_{\mu}^{-1}$ and $\Sigma_{\mu'}^{-1}$ are also updated, the performance decreases significantly.

\section{Results on a robotic manipulator}
\label{Results for a robotic manipulator}

This section evaluates the presented approach against the active inference controller (AIC) from \cite{pezzato2019novel}. We show that our approach outperforms the AIC from \cite{pezzato2019novel} at the task of manipulating different payloads, and setting different initial parameters for the variances and different values of $\tau$. More results (including time-varying target states) can be found in our other work \cite{2020UKRAS_baioumy, 2020ICRA_baioumy}. 

\subsection{Robustness of intial settings}

The AIC from \cite{pezzato2019novel} achieves adaptive control without an explicit dynamics model. However, it is sensitive to the initialization of its parameters. By slightly changing $\Sigma_{\mu}^{-1}$ for instance, the system suffers from severe oscillations and never converges to the target state.

If the AIC is tuned optimally ($\Sigma_{\Observation}^{-1} = 1.5I$, $\Sigma_{\Observation'}^{-1} = 0.5I$, $\Sigma_{\mu}^{-1} = 0.1I$ and $\Sigma_{\mu'}^{-1} = 0.5I$), this results in satisfactory behaviour. In this case, $I$ refers to the 7x7 identity matrix. However, if we vary $\Sigma_{\mu}^{-1} = 0.1I$ or to other values ($0.3I$ and $0.5I$), the performance gets considerably worse. In our approach, we update $\tau$, $\Sigma_{\Observation}$ and $\Sigma_{\Observation'}$ online to retune the controller. We ran the experiment for a pick-and-place task (as in \cite{pezzato2019novel}) for several values of $\Sigma_{\mu}^{-1}$  and recorded in Table \ref{table:varying simga} the Mean Absolute Error (MAE) defined as:
$$MAE  = \frac{1}{n_{t}} \sum\limits_{j=1}^{n_{t}} |\bm{\mu_d} - \bm{\mu}|.$$

When the AIC is properly tuned $\Sigma_{\mu}^{-1} = 0.1I$, the two cases have the same MAE (tuning does not matter since the initialization was optimal). However, when $\Sigma_{\mu}^{-1} = 0.3I$, the controller becomes does not converge to the target state (visualized in Figure \ref{fig:results for the real pands arm}) . The MAE increases to more than triple its value while in the case of tuning the hyperparameters, the MAE actually decreases. This is due to the fact that increasing $\Sigma_{\mu}^{-1}$ makes the controller more aggressive and when tuned, it does not oscillate and also has a slightly faster response.  In a similar fashion, results for changing the value of $\tau^{-1}$ are recorded in Table \ref{table:varying tau}. Again, the MAE is much lower when using our method.

\subsection{Results on adaptive behaviour}
For the last experiment, the robot carries varying payloads. All controllers are tuned to have identical performance in the no payload case. Subsequently, we test three different payloads: $m = 1kg$, $m = 2kg$ and $m = 3kg$ (max payload for the Panda arm). The MAE for these cases is recorded in the Table \ref{table:varying masses}.

\begin{table}[H]
\begin{center}
\begin{tabular}{@{}c|ccc@{}}
\multicolumn{1}{l|}{}      
& $\Sigma_{\mu}^{-1}$ = 0.1$I$ &  0.3$I$ &  0.5$I$ \\ \hline
No updates                                                            & 0.028                & 0.088                & 0.118               \\
Updating $\tau^{-1}$, $\Sigma_{\Observation}^{-1}$ and $\Sigma_{\Observation'}^{-1}$ & 0.028                & \textbf{0.025}       & \textbf{0.032}     
\end{tabular}
\caption{MAE for different values of $\Sigma_{\mu}^{-1}$.}
\label{table:varying simga}
\end{center}
\end{table}

\begin{table}[H]
\begin{center}
    
\begin{tabular}{@{}l|ll@{}}
                  & $\tau^{-1}$ = 2$I$     & $\tau^{-1}$ = 3$I$     \\ \hline
\multicolumn{1}{l|}{No updates}                    & 0.091          & 0.123          \\ 
\multicolumn{1}{l|}{Updating $\tau^{-1}$, $\Sigma_{\Observation}^{-1}$ and $\Sigma_{\Observation'}^{-1}$} & \textbf{0.025} & \textbf{0.032} \\ 
\end{tabular}
\caption{MAE for different values of $\tau^{-1}$.}
\label{table:varying tau}
\end{center}
\end{table}

\begin{table}[H]
\begin{center}
\begin{tabular}{@{}c|ccc@{}}
\multicolumn{1}{l|}{}      
& $m = 1kg$ & $m = 2kg$ & $m = 3kg$ \\ \hline
AIC no updates                                                            & 0.024                & 0.029                & 0.027               \\
AIC with updates (ours) & \textbf{0.020}                & \textbf{0.020}       & \textbf{{0.021}}    
\end{tabular}
\caption{Mean Absolute Error (MAE) for different payloads in case of updating hyperparameters and no updates.}
\label{table:varying masses}
\end{center}
\end{table}

\section{Discussion and future work}
\label{sec:discussion}
In the AIC, actions are not explicitly modelled. In Section~\ref{Actinve inference and free energy}, the generative model was selected to have the form $ p({\mathbf{\Observation}}, \bm{{\state}})$ which does not explicitly include any notion of an action $\mathbf{a}$. Thus to choose the action that minimizes free-energy, the chain rule was utilized (Equation \ref{eq:F_minimize_a}). Alternatively, the actions could be explicitly added in the generative model $p({\mathbf{\Observation}}, \bm{\state}, \mathbf{\action})$. When doing so, this problem can be efficiently solved using factor graphs \cite{loeliger2007factor, vander_laar_2019_lqg, vanderbroeck2019active}.

A key property of the AIC is the coupling between control and state-estimation. As shown, to perform any meaningful control, the state must be biased towards the target. Thus our belief about the true state is only accurate when the systems reaches the target. The coupling between estimation and control makes some quantities lose their physical meaning. For instance $\Sigma_{\Observation}$ does not represent the uncertainty of the joint position encoders anymore. It is rather a combination of the encoder's uncertainty and how far the target is form the current position. Future work will investigate this further.

\section{Conclusions}
\label{sec:conclusions}
In this paper we demonstrate how a minimizing a single quantity: variational free-energy, effective state-estimation, control and learning can be performed for robotic manipulator. Online estimation of relevant quantities can be achieved using gradient descent on the free-energy for each iteration of the controller. We introduce a temporal parameter and show that when $\tau$ approaches zero, the approach converts to a PID controller and if $\tau$ approaches $\infty$, it converts to a filter. We then demonstrated the effectiveness of the framework for a 7 DOF robotic arm and showed adaptability and robustness ourperforming previous work. We showed that out approach improves robustness, damps oscillations and adapts to different payloads.

\section*{ACKNOWLEDGMENT} \label{section:acknowledgment}
The authors thank Mees Vanderbroeck, Matias Mattamala and  Charlie Street for helpful comments and feedback.
This work was supported by UK Research and Innovation and EPSRC through the Robotics and Artificial Intelligence for Nuclear (RAIN), and Offshore Robotics for Certification of Assets (ORCA) hubs [EP/R026084/1, EP/R026173/1]. 

\bibliographystyle{unsrt}
\bibliography{sample.bib}

\end{document}